# MsMorph: An Unsupervised pyramid learning network for brain image registration

Jiaofen Nan[1], Gaodeng Fan[1], Kaifan Zhang[1], Chen Zhao[2], Fubao Zhu[1*], Weihua Zhou[3*]

1.School of Computer Science and Technology, Zhengzhou University of Light Industry, Zhengzhou, Henan Province, 450000 China

2. Department of Computer Science, Kennesaw State University, Marietta, GA, USA

3. Department of Applied Computing, Michigan Technological University, Houghton, MI, USA

Correspondence: fbzhu@zzuli.edu.cn; whzhou@mtu.edu

**Abstract**: In the field of medical image analysis, image registration is a crucial technique. Despite the numerous registration models that have been proposed, existing methods still fall short in terms of accuracy and interpretability. In this paper, we present MsMorph, a deep learning-based image registration framework aimed at mimicking the manual process of registering image pairs to achieve more similar deformations, where the registered image pairs exhibit consistency or similarity in features. By extracting the feature differences between image pairs across various aspects using gradients, the framework decodes semantic information at different scales and continuously compensates for the predicted deformation field, driving the optimization of parameters to significantly improve registration accuracy. The proposed method simulates the manual approach to registration, focusing on different regions of the image pairs and their neighborhoods to predict the deformation field between the two images, which provides strong interpretability. We compared several existing registration methods on two public brain MRI datasets, including LPBA and Mindboggle. The experimental results show that our method consistently outperforms state of the art in terms of metrics such as Dice score, Hausdorff distance, average symmetric surface distance, and non-Jacobian. The source code is publicly available at https://github.com/GaodengFan/MsMorph.

## I. INTRODUCTION

Deformable registration has always been a focus of attention in medical imaging, playing an important role in preoperative and postoperative guidance planning, as well as disease diagnosis [Sotiras et al.2013; Nan etal.2023; Haskins et al.2020]. For an image pair, consisting of a fixed image and a moving image, deformable registration predicts the nonlinear relationship between the two images, warps the moving image, and strives to make the warped image as similar or consistent as possible in spatial structure with the fixed image. In practice, doctors can deduce the correct location of organs based on this nonlinear relationship, thereby maximizing the accuracy and effectiveness of diagnosis or surgery.

Although traditional methods [Rueckert et al.1999; Avants et al.2008; Beg et al.2005; Nan et al.2023; Lowe 1999; Bay et al.2008; Thirion 1998; Heinrich et al.2015; Modat et al.2010] have achieved fairly high registration accuracy, computationally expensive and time-consuming. In recent years, with the development of hardware performance, neural network-based methods [Balakrishnan et al.2019; Liu et al.2022; Meng et al.2023; Chen et al.2022; Hu et al.2022; Wang et al.2024; Lv et al.2022; Jia et al.2022; Li et al.2022; Hoffmann et al.2021] have achieved performance superior to traditional methods, especially in terms of inference speed. A trained model can complete the registration of an image pair in about one second, greatly expanding the practical applicability of image registration in real-world environments. However, we pinpoint several shortcomings of previous neural network-based methods: (1) They often overlook the issue of inconsistent voxel distribution within images. A brain MRI scan encompasses multiple organs, and the voxel values for these different organs typically vary significantly; (2) They lack interpretability, functioning as 'black-box models' that only output the final deformation field, without explaining the reasons behind the alignment.

In this study, we propose MsMorph, an unsupervised deep learning framework for medical image registration, designed to address the aforementioned issues. We compare our proposed network with several existing registration methods on two public brain MRI datasets, and our method consistently outperformed the others. The main contributions are as follows:

(1) We propose a novel Unsupervised registration network for brain MRI images that effectively mimics the manual process of registering image pairs.

(2) Differential neighborhood information—the dissimilar regions of the image pair to be registered and their neighboring areas—can be utilized to derive the optimal transformation, with this information learned by the neural network based on the registered image pairs.

(3) Gradient information is introduced to calculate the variation between voxels, allowing for improved extraction of structural features and better fitting of network parameters.

(4) The model is interpretable, since the differential neighborhood information driving the alignment can be visualized.

## II. RELATED WORK

**Classical Methods**. Classical medical image registration methods can be categorized into two types: intensity-based methods and feature-based methods. These approaches utilize various similarity functions, deformation types, transformation constraints or regularization strategies, and optimization techniques [Wang et al.2023; Evan et al.2022]. Common intensity-based methods include Mean Squared Error (MSE), Normalized Cross-Correlation (NCC) [Avants et al.2009; Yoo et al.2009; Hermosillo et al.2002], and Mutual Information (MI) [Viola et al.1997; Gong et al.2013; Heinrich et al.2012]. The core approach is to define an energy function that measures the similarity between the two images and to iteratively search for a parameterized deformation field to optimize this energy function [Wang et al.2023]. Another class of feature-based registration paradigms often involves extracting keypoints using manual methods, contours, color information, segmented regions, or image intensities [Lowe 1999; Bay et al.2008; Toews et al.2013; Van et al.2005]. This method determines the optimal parameters by establishing corresponding relationships. Notable algorithms include Demons [Thirion 1998], which is based on optical flow concepts; the Large Deformation Diffeomorphic Metric Mapping (LDDMM) [Beg et al.2005], which employs time-independent velocity fields; the improved SyN algorithm based on LDDMM; NiftyReg [Modat et al.2010], which integrates rigid registration, affine alignment, and nonlinear registration; and scale-invariant keypoint matching algorithms like SIFT [Lowe 1999] and SURF [Bay et al.2008]. While these methods have experienced continuous advancements in mathematical models, optimization algorithms, and similarity metrics, they often suffer from drawbacks such as computationally expensive and time-consuming.

**Learning-based Methods**. Learning-based methods have become a prominent trend in medical image registration research, effectively addressing the time-consuming nature of traditional methods and significantly improving usability in real-time applications. [Balakrishnan et al.2019] introduced VoxelMorph, an unsupervised image registration method based on U-Net [Ronneberger et al.2015], which marked a pivotal shift from traditional approaches to deep learning methods. Since then, research has surged, with models like Vit-V-Net [Chen et al.2021], which introduced the Transformer [Vaswani 2017], and later TransMorph [Chen et al.2022], which integrated the Swin Transformer [Liu et al.2021] to replace convolutions as the encoder in a U-Net architecture, offering multiple variants. As research advanced, continuous deformation approaches gained increasing attention, primarily divided into pyramid and cascaded models. The pyramid approach involves multi-scale predictions, with each layer predicting a deformation field, and these predictions are ultimately fused into an optimal solution. In contrast, the cascaded approach recursively predicts deformation fields within the network, progressively refining the solution. Both approaches focus on continuous deformation. VTN [Zhao et al.2019a] and RCN [Zhao et al.2019b] were among the earliest deep learning networks to utilize continuous deformation for medical image registration. Dual-PRNet [X et al.2019] introduced a dual-stream pyramid registration network that explicitly enhanced model correlation estimation. PR++[ Kang et al.2022] built upon the PR module by adding 3D correlation layers and residual connections, directly utilizing the deformation field to transform moving feature maps. ModeT [Wang et al.2023] incorporated attention mechanisms at each layer of the pyramid, along with channel competition at specific layers to drive the solution of reasonable deformations. A recent study, RDP [Wang et al.2024], utilized residuals as the backbone for feature extraction, iteratively predicting residual deformation fields at each decoder layer, thereby enhancing semantic information across different resolution levels and within level loops to maximize the accuracy of the deformation field.

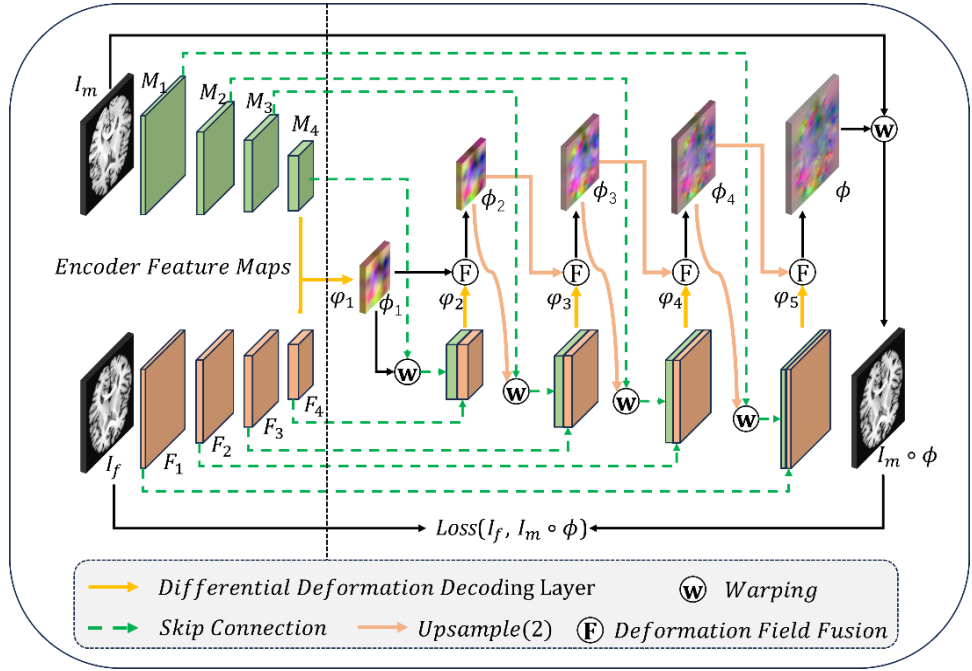

Figure 1: Proposed framework. The 3D images $I_f$ and $I_m$ extract hierarchical features $F_1$ - $F_4$ and $M_1$ - $M_4$ through a dual-stream encoder with shared weights. The decoder, consisting of Differential Deformation Decoding Layers and Deformation Field Fusion blocks, predicts the deformation field $\phi$ from coarse to fine, and computes the loss between $I_f$ and $I_m \circ \phi$ to guide training. $I_m \circ \phi$ represents the deformation of $I_m$ with the final derived deformation field $\phi$, and the Warping operation is performed by the spatial transformer network (STN)[ Jaderberg et al.2015].

## III. METHOD

The proposed deep learning registration network framework is illustrated in Fig. 1. The work seeks to mimic the manual process of image pair registration, which involves iterative comparison and warping. Specifically, by comparing the differences between image pairs and applying deformations based on these differences to minimize the disparity, the process is repeated to produce an increasingly accurate deformation field. It is clear that this is a continuous deformation process, involving multiple rounds of difference comparison and deformation estimation, where the final deformation field is the result of combining several incremental deformations. The core idea can be represented by the following formula:

$$I_f \sim Warped = I_m \circ \phi = I_m \circ \phi_1 \circ \phi_2 \circ \ldots \circ \phi_i \circ \ldots \circ \phi_n \tag{1}$$

where 'o' represents the deformation operation, which is typically implemented by the Spatial Transformation Network (STN) [Jaderberg et al.2015]. Specifically, for any voxel $p$, its position is $p'$ in the moving image $I_m$. Since $p'$ may not be an integer, interpolation methods (such as Equation (2)) are required to compute the voxel value in the deformed image:

$$I_m \circ \phi(p) = \sum_{q=Z(p')} I_m(q) \prod_{d \in \{x,y,z\}} (1 - |p'_d - q_d|) \tag{2}$$

where $Z(p')$ represents the nearest integer coordinates to $p'$, and $d$ iterates over the dimensions of the entire volume domain $\Omega$.

Based on the above ideas, the work exploits the scale invariance of images to extract features at multiple levels within a pyramid structure, forming the registration network shown in Fig. 1. By comparing feature differences across these levels, the network computes multi-level deformation quantities, which are subsequently combined into the final deformation field. This process effectively replicates the manual registration of image pairs. Compared to methods that rely solely on the original image, this multi-resolution strategy provides several advantages, including a larger convergence radius (or capture range) and increased robustness against local optima [Lipson et al.2021]. The network is built on a U-Net model with a dual-stream encoder structure. The decoder primarily consists of differential deformation decoding layers and deformation fusion blocks. At four resolution levels, the network calculates the differences between image pairs, capturing the maximum possible information to estimate the deformation field.

### A. Encoder architecture

The encoder serves the purpose of feature extraction and adopts a dual-stream pyramid structure with parameter sharing. Structural information within images is crucial for registration. To extract this spatial information, spatial attention mechanisms are referenced, employing average pooling to capture features and learn this spatial information across multiple scales. The registration of medical images primarily addresses the discrepancies between image pairs. Therefore, in this study, moving image $I_m$ and the fixed image $I_f$ are not directly concatenated. Instead, both images are input separately into parameter-sharing encoders for feature extraction. Distinct features are extracted from different images using the same parameters, which enhances the clarity of dissimilar features. This approach facilitates the modeling of relational mappings based on these differences.

The encoder consists of four layers, as shown in Fig. 2, designed to extract features at different resolutions. $I_f$ and $I_m$ are input to their respective encoders, then the feature maps $F_1$ and $M_1$ are generated after the 3×3×3 convolution ( Conv ), Instance Normalization ( InsNorm ), and a nonlinear activation function (LRelu). In the following three layers, the feature maps from the previous layer are first downsampled by an Average pooling layer ( AvgPool ), followed by two Conv, an InsNorm, and an LReLU. This process generates the corresponding feature maps, resulting in two sets of feature maps $F_1$, $F_2$, $F_3$, $F_4$ and $M_1$, $M_2$, $M_3$, $M_4$. It is important to note that only the first layer does not perform downsampling, resulting in the volumes of $F_1$ and $M_1$ being identical to those of the original input images. In the subsequent three layers, however, the volume is reduced by a factor of $(1/2)^3$ at each layer. Furthermore, the channel count follows the pattern $C \times 2^{l-1}$, where $l$=1,2,3,4.

### B. MsMorph

Given a pair of images, including the moving image $I_m$ and the fixed image $I_f$, both images are fed into the encoder to obtain their corresponding multi-scale feature representations $M_l$ and $F_l$, where $l$=1,2,3,4. The proposed network contains a total of four decoding layers, and through the learning of multi-level features, it iteratively estimates a deformation field that progressively approaches the target, achieving fine registration from coarse to fine. To compute the deformation field $\phi_l$ at layer $l$, the Differential Deformation Decoding Layer(*DDDL*) takes the moving and fixed feature maps as input to estimate sub deformation field $\varphi_l$, as illustrated in Fig. 2. Specifically, the *DDDL* learns the feature differences between the input image features and multiplies the learned feature weights with the moving and fixed feature maps. These results are then aggregated and passed through the *Deformation estimation (Def)* to obtain sub deformation field $\varphi_l$, see Section III.C. Finally, the newly predicted sub deformation field $\varphi_l$ is combined with previous deformation field $\phi_{l-1}$ using the synthesis method from [Zhao et al.2019a] to deformation field $\phi_s$ (see Equation (7)). This process involves residual connections instead of merely accumulating deformation fields. This strategy enhances the network's expressiveness, allowing it to fit parameters more effectively.

For a pair of images $I_m$ and $I_f$, the encoder produces two sets of feature maps:

$$F_1, F_2, F_3, F_4 = Encoder(I_f) \quad (3)$$
$$M_1, M_2, M_3, M_4 = Encoder(I_m) \quad (4)$$

In the decoding layer of the 4-th layer, the feature maps $M_4$ and $F_4$ are input into *DDDL* (see Section III.C) to generate the first sub deformation field $\varphi_1$, which, since no new deformation field has been produced yet, can also serve as the initial deformation field $\phi_1$:

$$\phi_1 = \varphi_1 = DDDL(F_4, M_4) \quad (5)$$

After obtaining $\phi_1$, $F_4$ and $M_4 \circ \phi_1$ as input to estimate the sub deformation field $\varphi_2$. Both $\phi_1$ and $\varphi_2$ are then input into the Deformation Field Fusion(*DFF*) to merge the deformation fields, resulting in $\phi_2$. After obtaining the upsampled deformation field, the following process is performed:

$$\varphi_s = DDDL(F_l, M_l \circ \phi_{s-1}) \quad (6)$$
$$\phi_s = DFF(\phi_{s-1}, \varphi_s) = \phi_{s-1} \circ \varphi_s + \varphi_s \quad (7)$$
$$\phi_s = Up(2 \cdot \phi_{s-1}) \quad (8)$$

in this case, $s = 2,3,4,5$, and $l = 6 - s$, since $l$ starts decreasing from the 4-th decoding layer, while $s$ begins counting from 2. Consequently, the sum of $s$ and $l$ always equals 6. And $Up$ means trilinear interpolation upsampling operation. In such a way, it enables the transmission of high-level semantics between layers. After obtaining the deformation field $\phi_5$ at $l = 1$ (the first decoding layer), there is no longer a need to calculate $\phi_5$ using Equation (8). At this stage, $\phi_5$ serves as the final deformation field.

### C. Differential Deformation Decoding Layer

The purpose of image registration is to align two images so that they are as spatially congruent as possible. The manual registration process typically initiates with a comparative analysis of the images. The goal is to identify their differing regions and then enhance the similarity of the image pairs by dragging or stretching along the edges of these regions. This meticulous adjustment ensures that the images are rendered with the highest degree of spatial similarity, achieving the desired registration effect.

In deep learning registration networks, each voxel in the feature map contains rich semantic information, encompassing various potential deformations. To tackle this challenge, a sophisticated Differential Deformation Decoding Layer, as illustrated in Fig. 2, is employed to analyze deformation patterns at low-resolution levels.

When two identical images are processed through the same encoder, the extracted features are identical. Conversely, if the two images differ, the features obtained from the same encoder will also differ. Therefore, after registration, the features should be similar, if not identical, to some extent.

Gradients are highly sensitive to variations between voxels. A brain image comprises numerous organs, each typically exhibiting distinct pixel values. Gradients can effectively detect these variations, enabling the identification of regions where pixel changes occur. Consequently, gradients are introduced to capture areas of pixel variation within the image, facilitating the extraction of features that enhance the modeling of the registration network.

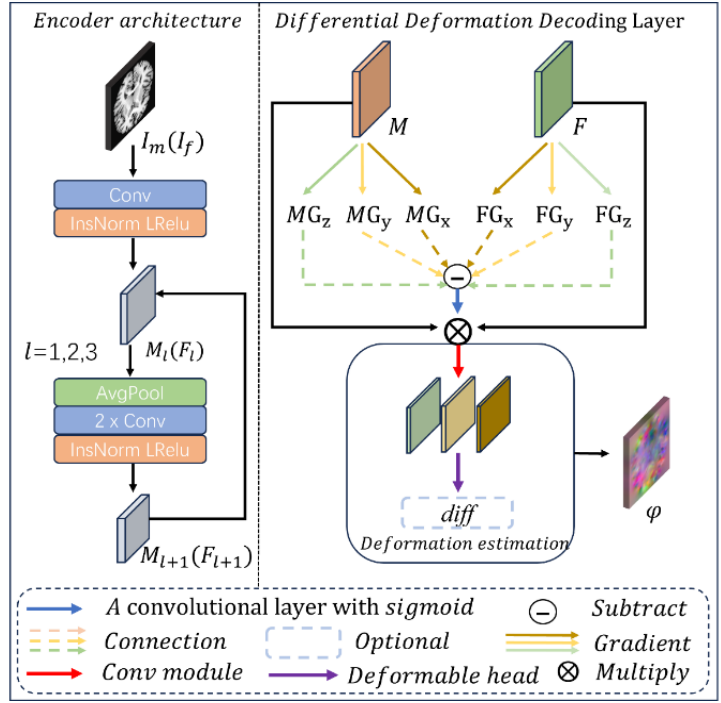

Figure 2: Details of the Encoder architecture and the Differential Deformation Decoding Layer.

The $DDDL$ takes $F_l$ and $M_l$ as inputs, and calculates their gradient information in the x, y and z directions, resulting in $FG_x$, $FG_y$, $FG_z$, for the $I_f$ and $MG_x$, $MG_y$, $MG_z$ for the $I_m$. Considering that the boundaries between similar and dissimilar regions also impact the registration outcome, convolutional layer is employed to extract neighborhood information from the dissimilar regions. This information is then processed through a sigmoid function to yield a weighted feature representation that influences the registration process, as illustrated in Fig. 2.

$$W_{i(direction)} = sigmoid\left(Conv\left(FG_{i(direction)} - MG_{i(direction)}\right)\right) \qquad (9)$$

$$\varphi_{i(direction)} = Def\left(F_i \times W_{i(direction)}, M_i \times W_{i(direction)}\right) \qquad (10)$$

The obtained weights are multiplied with the input features, and after passing through the $Def$ (deformation estimation block), the field $\varphi_{i(direction)}$ for a specific direction is computed. Ultimately, the deformation fields from the three directions are concatenated to sub deformation field $\varphi$. The diffeomorphism layer ($diff$) is optional and serves to ensure diffeomorphism, which from [Wang et al.2024]. The proposed method closely resembles human processing approaches.

D. Loss Function

By minimizing the difference between $I_f$ and $I_m \circ \phi$, the network is incentivized to learn optimal parameters. The Normalized Cross-Correlation (NCC), which is robust to intensity variations across scans and datasets[Avants et al.2008], is employed to measure similarity. A higher NCC indicates greater similarity between the two images, leading to the similarity loss defined as $L_{sim}$ = - NCC ($I_f, I_m \circ \phi$). The regularization term $L_{reg}$ constrains the deformation field $\theta$ to ensure a smooth deformation by computing the $L_2$-norm of its gradient. Thus, for each voxel $p$ in the entire volumetric domain $\Omega$, the formulation is as follows:

$$L_{reg}(\phi) = \sum_{p \in \Omega} \|\nabla \phi(p)\|^2 \qquad (11)$$

Therefore, the loss function is:

$$\underset{\phi}{argmin} L_{sim}(I_f, I_m \circ \phi) + \lambda L_{reg}(\phi) \qquad (12)$$

where λ is the weight of the regularization term.

## IV. EXPERIMENTS

### A. Datasets

The proposed method was thoroughly validated on two publicly available brain MRI datasets, including LPBA [Shattuck et al.2008]and Mindboggle [Klein et al.2012].

The LPBA dataset contains T1-weighted 3D brain MRI data from 40 human volunteers, along with 54 manually labeled anatomical structures (ROIs) for each volume. All images have been rigidly pre-aligned to the MNI305, the rigidly pre-alignment refers to affine transformation. The volumes were center-cropped to a size of 160 × 192 × 160 (1 mm × 1 mm × 1 mm). The data was randomly divided into two groups: 30 volumes for the training set and 10 volumes for the testing set. Consequently, the training set consists of 30 × 29 pairs for registration, while the testing set contains 10 × 9 pairs.

We selected NKI-RS-22, NKI-TRT-20, and OASIS-TRT-20 from the Mindboggle dataset, which includes 62 manually labeled ROIs for each volume, with all images pre-aligned to the MNI152. Similarly, all images were center-cropped to a size of 160 × 192 × 160 (1mm × 1mm × 1mm). The NKI-RS-22 and NKI-TRT-20 datasets were combined to form a training set of 42 volumes (42 × 41 pairs), while 20 images from OASIS-TRT-20 (20 × 19 pairs) were used as the testing set.

### B. Evaluation Metrics

To objectively assess the registration performance of the proposed method, a variety of metrics were utilized. These included the average Dice Similarity Coefficient (DSC) [Dice 1945], the 95th percentile Hausdorff distance (HD95) [Huttenlocher et al.1993], the Average Symmetric Surface Distance (ASSD) [Taha et al.2015], and the percentage of voxels with non-positive Jacobian determinants ($\%|J_\phi| \leq 0$) [Ashburner 2007]. DSC, HD95, and ASSD are metrics that quantify the overlap of the corresponding regions of interest (ROIs). The metric Jacobian assesses the diffeomorphic properties of the deformation field. Additionally, optimal registration results are indicated by a higher DSC and lower values for HD95, ASSD, and Jacobian.

### C. Implementation Details

The proposed method was implemented in PyTorch [Paszke et al.2019] and executed on a workstation equipped with an NVIDIA V100 GPU with 32GB of memory. The batch size was set to 1, and training was conducted for 30 epochs with an initial learning rate ($lr_{init}$) of 0.0001, which decayed dynamically as training progressed (see Equation (13) ). The regularization parameter λ in Equation (12) was set to 1, and the Adam optimizer was employed for fine-tuning the model weights. Additionally, two variants of the proposed method, *msmorph* and *msmorph_diff*, were trained, corresponding to the presence or absence of the *diff* layer.

$$lr_e = lr_{init} \times \left(\frac{30 - e}{30}\right)^{0.9} \quad (13)$$

Where $e \in [0,29]$ and $lr_e$ represents the learning rate at epoch = e.

### D. Comparison Methods

To validate the superiority of the proposed method, a comparison was made with several state-of-the-art registration methods: (1) SyN [Avants et al.2008]: A method from the Advanced Normalizing Tools (ANTs), commonly used in the medical registration field, which employs mean squares as the optimization target and sets the iterative vector to (160, 80, 40). (2) VM (VoxelMorph) [Balakrishnan et al.2019]: A widely popular U-Net structured network model. (3) TM (TransMorph) [Chen et al.2022]: A U-Net architecture utilizing a Swin Transformer as the encoder and CNN as the decoder. (4) I2G (im2grid) [Liu et al.2022]: A pyramid structure network model that incorporates neighborhood attention. (5) RDN [Hu et al.2022]: A pyramid network that recursively decomposes the deformation field. (6) NiceNet [Meng et al.2023]: A non-iterative pyramid network for joint affine and deformable image registration. (7) PR++ [Kang et al.2022]: A pyramid registration network using 3D correlation layer. (8) RDP [Wang et al.2024]: A pyramid registration network that recursively decomposes the deformation field using a residual network as the feature extraction backbone.

To ensure an objective comparison, existing public implementations were employed, and recommended parameters were selected for training.

## E. Results and Discussion

### 1) Comparison with State of Arts:

The experimental results on the LPBA and Mindboggle datasets are summarized in Tables 1 and 2. The proposed method demonstrates superior performance in terms of DSC, HD95, ASSD, indicating that it can accurately predict the registration deformation field. Additionally, the proposed method aligns the edge regions of initial labels more precisely than other methods (see Fig. 3). This is particularly evident in the LPBA dataset, where both variants outperform all comparative methods across the four metrics. In the Mindboggle dataset, both variants exhibit slight inferiority in the negative Jacobian determinant compared to the classical method SyN; however, MsMorph_diff still surpasses other methods, emerging as the second-best performer. This confirms that the introduction of the differential diffeomorphism layer facilitates more reasonable deformations.

Among all comparative methods, single-stage U-Net networks such as VM and TM achieve registration accuracy comparable to classical methods but fall short of pyramid-based approaches. The second-best method, RDP, benefits from multiple recursive iterations that continuously mitigate prediction errors, enhancing the accuracy of the deformation field. Another effective method, NiceNet, benefits from Swin Transformer and the continuous fusion of transmitted semantic information. Notably, despite not employing multiple recursive iterations or a Transformer architecture, the proposed method still achieves the best results, attributed to its ability to mimic human cognitive processes during registration, with a particular focus on dissimilar regions.

Table 1: Results of different registration models on the LPBA (54 ROIs) dataset. The bolded evaluation values indicate the best performance.

| Model | DSC(%) ↑ | $\%|J_\phi| \leq 0$ ↓ | HD95 ↓ | ASSD ↓ |
|---|---|---|---|---|
| Affine | 53.7± 4.8 | - | 7.43±0.78 | 2.74±0.33 |
| SyN | 70.4±1.8 | <0.002% | 5.82±0.51 | 1.72±0.12 |
| VM | 67.1±2.8 | <0.8% | 6.37±0.64 | 1.89±0.20 |
| TM | 68.5±2.8 | <0.6% | 6.27±0.65 | 1.82±0.20 |
| I2G | 70.6±1.5 | <0.01% | 5.74±0.49 | 1.66±0.12 |
| RDN | 71.5±1.6 | <0.3% | 5.69±0.48 | 1.63±0.12 |
| NiceNet | 72.8 ±1.4 | <0.5% | 5.49±0.45 | 1.55±0.10 |
| PR++ | 69.6±2.3 | <0.1% | 6.11±0.61 | 1.76±0.17 |
| RDP | 73.1±1.7 | <0.0008% | 5.53±0.50 | 1.55±0.12 |
| MsMorph | **73.4±1.4** | <0.1% | 5.44±0.47 | 1.53±0.10 |
| MsMorph_diff | 73.4±1.5 | **<0.00009%** | **5.44±0.46** | **1.52±0.10** |

Table 2: Results of different registration models on the Mindboggle (62 ROIs) dataset. The bolded evaluation values indicate the best performance.

| Model | DSC(%) ↑ | %$|J_\phi| \leq 0$ ↓ | HD95 ↓ | ASSD ↓ |
|---|---|---|---|---|
| Affine | 31.7±2.4 | - | 7.12±0.61 | 2.26±0.24 |
| SyN | 56.5±1.4 | **<0.00005%** | 5.69±0.41 | 1.40±0.10 |
| Vm | 56.5±2.4 | <1.2% | 6.39±0.55 | 1.56±0.18 |
| Tm | 59.3±2.3 | <1% | 6.12±0.54 | 1.45±0.16 |
| I2G | 59.3±1.5 | <0.02% | 5.71±0.39 | 1.35±0.11 |
| RDN | 61.6±1.4 | <0.004% | 5.46±0.43 | 1.26±0.10 |
| NiceNet | 64.6±1.3 | <0.9% | 5.39±0.40 | 1.21±0.09 |
| PR++ | 58.7±2.1 | <0.5% | 6.17±0.53 | 1.46±0.16 |
| RDP | **64.8±1.3** | <0.01% | 5.47±0.42 | 1.22±0.10 |
| MsMorph | 64.6±1.3 | <0.3% | 5.39±0.40 | 1.21±0.09 |
| MsMorph_diff | **64.8±1.3** | <0.004% | **5.39±0.39** | **1.20±0.09** |

We visualized the registration results for all methods on the two datasets, as shown in Fig. 3, where it is evident that the proposed method generates more accurately registered images. Additionally, several regions from both the left and right hemispheres were combined for quantitative analysis, with results presented in Fig. 4 and 5. Fig.4 displays the DSC distribution for 11 brain regions in the LPBA dataset, including the frontal lobe, parietal lobe, temporal lobe, occipital lobe, cerebellum, brainstem, orbital frontal lobe, cingulate gyrus, caudate nucleus, putamen, and hippocampus. Fig. 5 illustrates the DSC distribution for 13 brain regions in the Mindboggle dataset, including the frontal lobe, parietal lobe, occipital lobe, hippocampus, orbital frontal lobe, cingulate gyrus, caudate nucleus, cerebrospinal fluid, third ventricle, fourth ventricle, and brainstem. The proposed method significantly outperforms all methods across all ROIs.

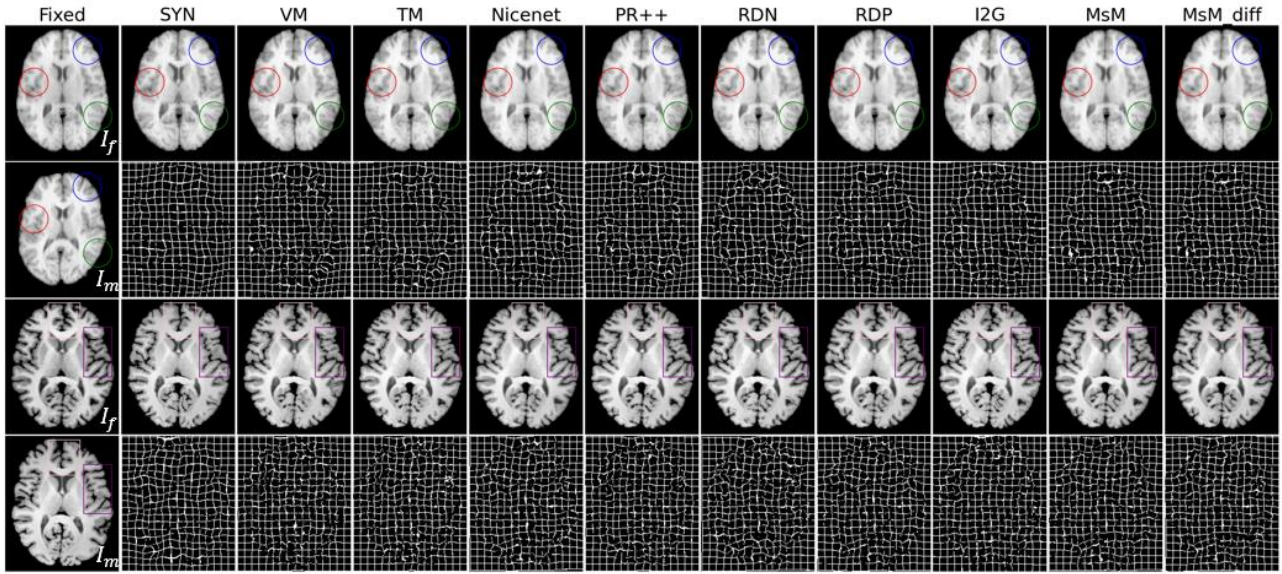

Figure 3: Example results of different registration models on two datasets. Rows 1 and 3 show the registered images, while rows 2 and 4 visualize the corresponding deformation fields under a grid.

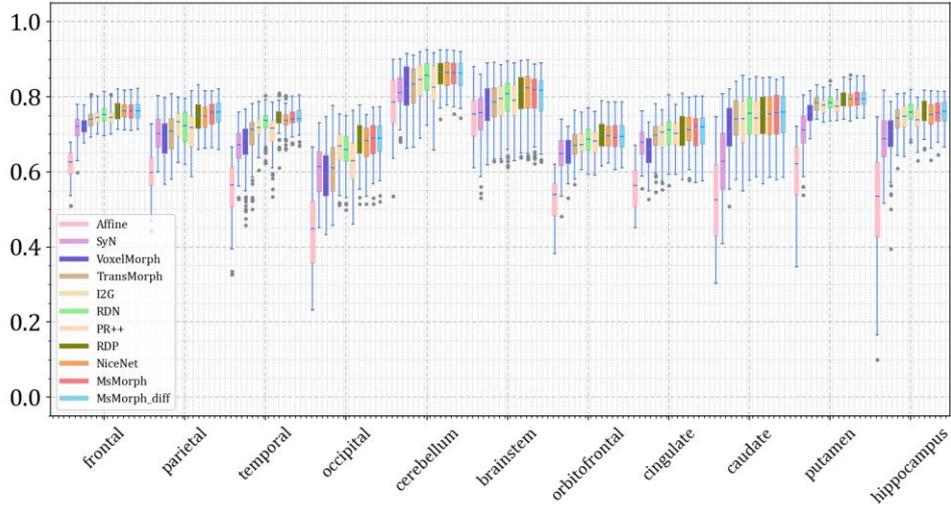

Figure 4: Box plots of the DSC scores for the proposed MsMorph and MsMorph_diff methods, as well as existing image registration techniques, are presented across 11 regions in the LPBA dataset.

## 2) Analysis of the Continuous Deformation

The proposed registration network employs a multi-level decomposition of the deformation field. Fig. 6 and 7 illustrate the continuous deformation process for a pair of images from both datasets under the MsMorph and MsMorph_diff frameworks. Notably, since the lowest level predicts twice, $\phi_1$ and $\phi_2$ appear in the same color. The figures clearly show that as the deformation field gains more information, the details in the distorted image increasingly resemble those in the fixed image. Using the DSC metric, we evaluated $\phi_{1-5}$ on both datasets, with results summarized in Table 3, confirming the gradual refinement of the estimated deformation field. It is evident that as resolution increases, detailed features accumulate, resulting in a more refined deformation field. Additionally, it was noted that the DSC results for $\phi_1$ and $\phi_2$ after deformation showed little difference, while registration performance across subsequent layers varied significantly. This suggests that increasing number of pyramid levels may diminish its impact on registration effectiveness beyond a certain point, a conclusion that will be further investigated in section IV.E.6.

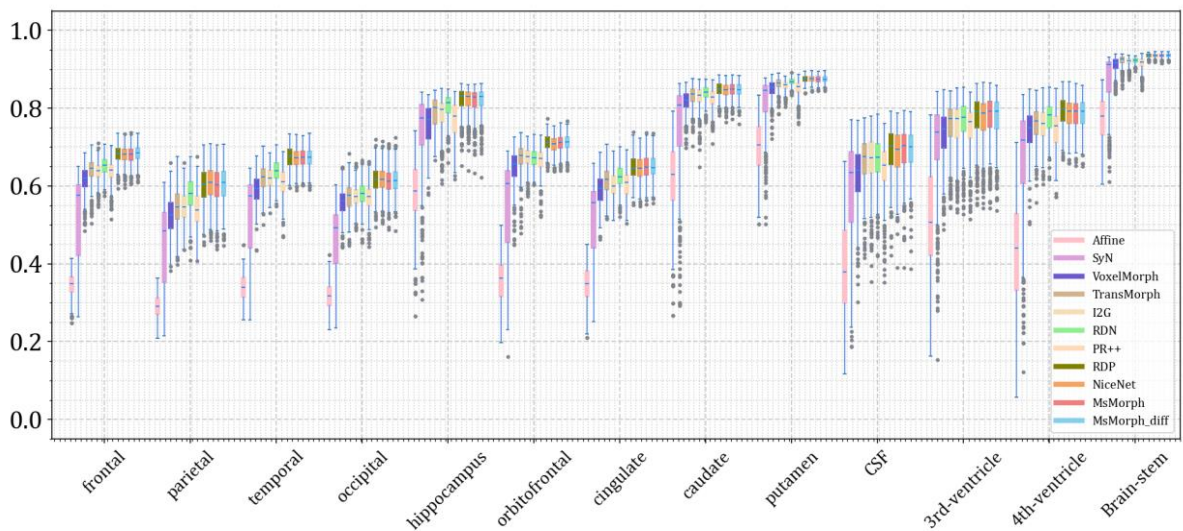

Figure 5: Box plots of the DSC scores for the proposed MsMorph and MsMorph_diff methods, as well as existing image registration techniques, are presented across 13 regions in the Mindboggle dataset.

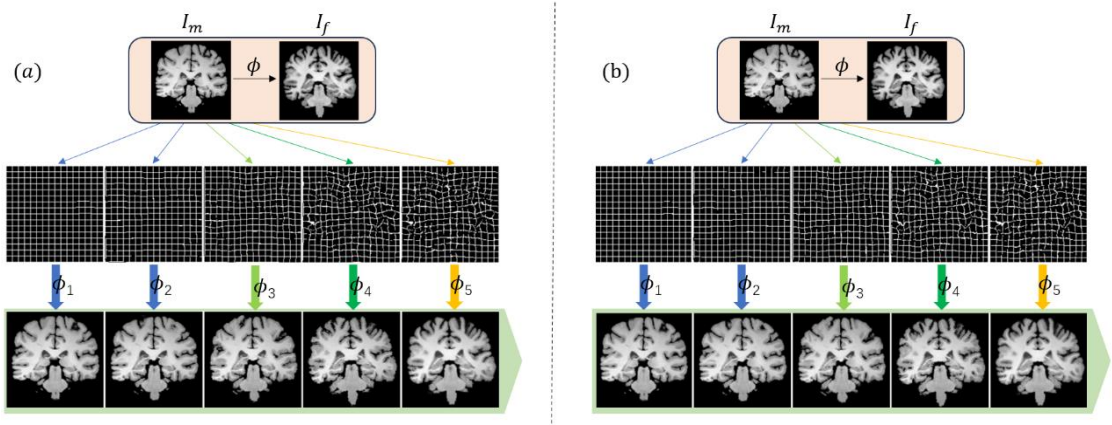

Figure 6: A continuous deformation process for the same pair of images on the Mindboggle dataset for the two variants. (a) Deformation process driven by MsMorph; (b) Deformation process driven by MsMorph_diff. The first row presents $I_m$ alongside $I_f$, while the second row displays the predicted deformation fields at each pyramid level.

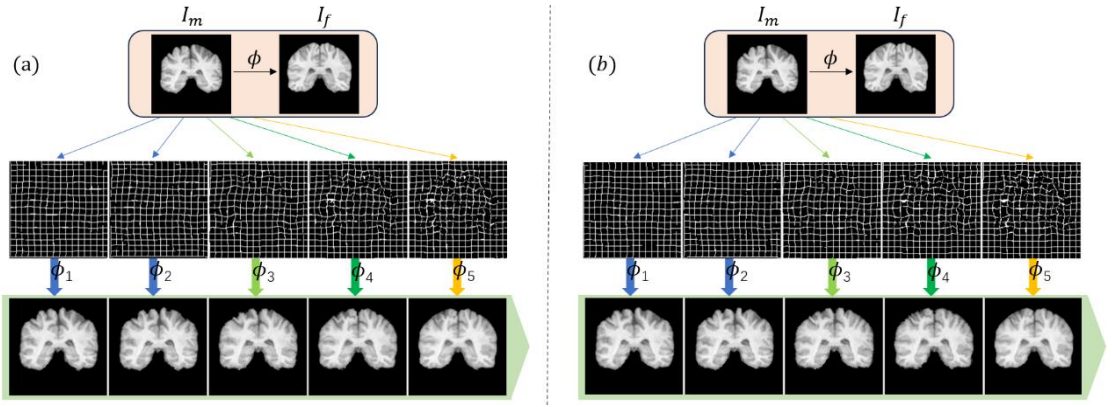

Figure 7: A continuous deformation process for the same pair of images on the LPBA dataset for the two variants. (a) Deformation process driven by MsMorph; (b) Deformation process driven by MsMorph_diff. The first row presents $I_m$ alongside $I_f$, while the second row displays the predicted deformation fields at each pyramid level.

Table 3: DSC (%) results for different deformation fields ($\phi_1$ - $\phi_5$) in the continuous deformation process.

| Datasets | Model | $\phi_1$ | $\phi_2$ | $\phi_3$ | $\phi_4$ | $\phi_5$ |
|---|---|---|---|---|---|---|
| LPBA | MsMorph | 54.4±4.4 | 55.2±4.1 | 58.8±3.8 | 64.6±2.7 | 73.3±1.5 |
|  | MsMorph_diff | 54.5±4.4 | 55.2±4.1 | 58.8±3.8 | 64.4±2.7 | 73.4±1.5 |
| Mindboggle | MsMorph | 31.9±2.3 | 34.3±2.1 | 39.9±1.9 | 59.8±1.3 | 64.6±1.3 |
|  | MsMorph_diff | 32.0±2.3 | 34.2±2.0 | 39.7±1.8 | 59.3±1.2 | 64.8±1.3 |

### 3) Analysis of the differential neighborhood visualization

We visualized the differential neighborhood, with Fig. 8 and 9 showcasing examples from the registration process on the LPBA and Mindboggle datasets. Since $\phi_5$ represents the final deformation field, the model does not compute the differential neighborhood after the deformation of $\phi_5$. Initially, significant differences are observed between the two images; however, as the registration process advances from coarse to fine, these differences gradually diminish. This indicates that the model effectively transitions from addressing large-scale differences to focusing on specific detail discrepancies, thereby facilitating the refinement of the deformation field and ensuring greater accuracy in similarity.

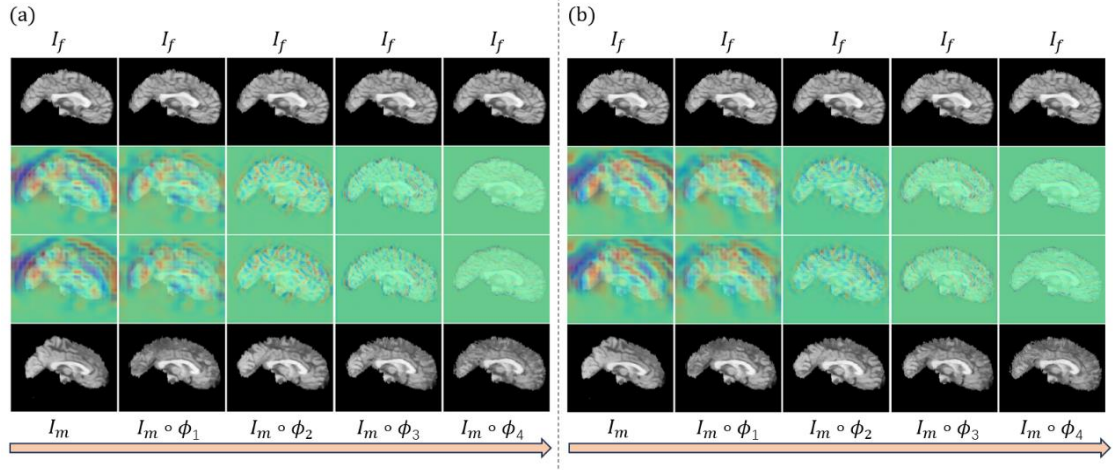

Figure 8: An example of differential neighborhood visualization during a representative continuous deformation process on the LPBA dataset: (a) is generated by MsMorph; (b) is generated by MsMorph_diff. The first row shows the fixed image, while the last row illustrates the continuous deformation process of the moving image. The second and third rows display the differential neighborhood features overlaid on the fixed and moving images, respectively.

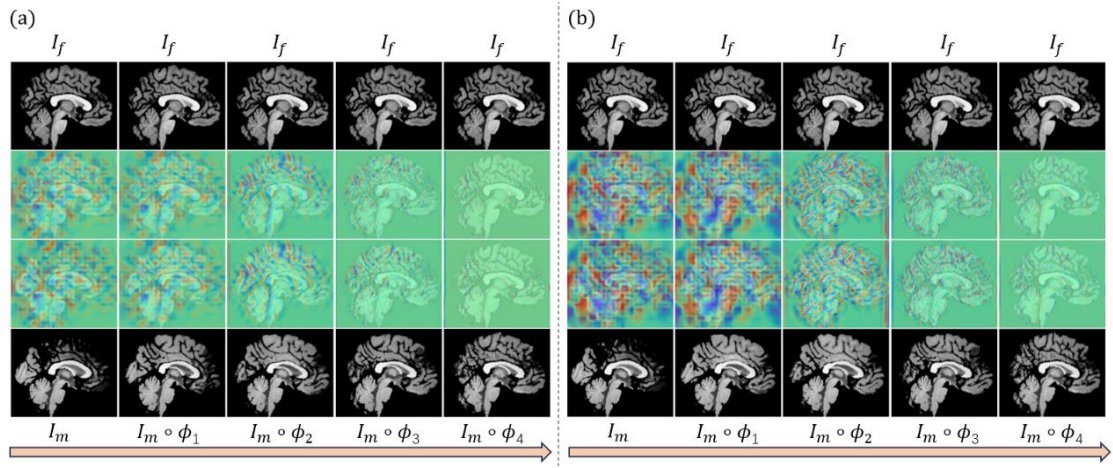

Figure 9: An example of differential neighborhood visualization during a representative continuous deformation process on the Mindboggle dataset: (a) is generated by MsMorph; (b) is generated by MsMorph_diff. The first row shows the fixed image, while the last row illustrates the continuous deformation process of the moving image. The second and third rows display the differential neighborhood features overlaid on the fixed and moving images, respectively.

4)   Analysis of the inference performance

Table 4 presents the average inference time and parameter count for different methods when processing each pair of images across the two datasets. Notably, SyN operates on a CPU, resulting in longer processing times compared to other GPU-based methods. VM exhibits the shortest inference time and the smallest parameter count among all methods, making it the least resource-intensive in practical applications, although it yields the poorest results. The parameter count of the two variants of our proposed method ranks just below TM and RDN. However, compared to these two methods, our approach significantly outperforms them in terms of similarity and spatial smoothness across the four metrics (see Tables 1 and 2). Moreover, the proposed method can run on Nvidia's entry-level GPU(RTX 4060) , making it suitable for practical applications due to its timely inference and enhanced accuracy.

Table 4: Average Inference Time and Model Parameter Count for Different Methods on Two Datasets.

| Model | Inference time(s) | | Parameters(M) |
|---|---|---|---|
| | LPBA | Mindboggle | |
| SyN | 159.956±18.455 | 167.963±11.453 | - |
| VM | 0.245±0.298 | 0.221±0.146 | 0.30 |
| TM | 0.521±0.382 | 0.400±0.173 | 46.77 |
| I2G | 0.848±0.497 | 0.898±0.277 | 0.89 |
| RDN | 1.121±0.362 | 1.032±0.280 | 28.65 |
| NiceNet | 0.820±0.354 | 0.773±0.187 | 5.71 |
| PR++ | 1.218±0.362 | 1.012±0.286 | 1.24 |
| RDP | 1.581±0.431 | 1.552±0.338 | 8.92 |
| Our | 0.969±0.382 | 0.931±0.207 | 14.55 |
| Our_diff | 1.145±0.530 | 1.233±0.363 | 14.55 |

5) Analysis of the Encoder performance

Table 5: DSC(%) Performance of Different Encoders on LPBA and Mindboggle.

| References | Encoder | DSC(%)↑ | |
|---|---|---|---|
| | | LPBA | Mindboggle |
| Liu et al.2022 | APolNLEncoder | 70.6±1.5 | 59.3±1.5 |
| Meng et al.2023 | APolNLEncoder | 72.8 ±1.4 | 64.6±1.3 |
| Hu et al.2022 | ConvEncoder-4 | 71.5±1.6 | 61.6±1.4 |
| Wang et al.2024 | ReEncoder | 73.1±1.7 | **64.8±1.3** |
| Kang et al.2022 | ConvEncoder-5 | 69.6±2.3 | 58.7±2.1 |
| MsMorph | ReEncoder | 72.9±1.7 | 64.3±1.4 |
| MSMorph(ours) | APolEncoder | **73.4±1.4** | 64.6±1.3 |

Table 5 presents the performance results of several pyramid methods, with the second column detailing the encoders used in the corresponding references, named based on their characteristics. Both APolNLEncoder and APolEncoder utilize AvgPool for downsampling; however, the former adds InsNorm and LReLU after each convolution. An excessive number of InsNorm layers can increase computational overhead and may overly emphasize detailed information while neglecting broader contextual cues. Consequently, APolEncoder reduces the number of normalizations and non-linear activations, incorporating them only at the end of each layer. The data in the table indicate that the results of MsMorph(ours) differ significantly from those in [Liu et al.2022]. In contrast, MsMorph(ours) shows a significant difference in registration results compared to [Meng et al., 2023] on the LPBA dataset, while the differences in registration results on the Mindboggle dataset are not as pronounced. Proposed method excels in capturing these differences, enabling more precise registration. ConvEncoder-l generates feature maps of varying resolutions and employs a convolutional approach with a stride of 2 for downsampling, which somewhat mitigates the information loss typically associated with downsampling. ReEncoder builds upon this by adding residual connections, creating a residual feature extraction backbone that enhances its receptive field. In experiments with MsMorph using ReEncoder as the encoder, as shown in Table 5, our encoder achieves a 0.5% and 0.3% improvement in DSC compared to the residual encoder in [Wang et al.2024] across the two datasets, demonstrating the effectiveness of APolEncoder. Additionally, our method primarily calculates the deformation field based on the differences between image pairs. Images often contain outlier points (noise) that can affect registration results. Since AvgPool averages all pixels within each pooling region, it exhibits a degree of noise resilience, making APolEncoder particularly suitable for our method.

6) Analysis of the Impact of Pyramid Levels

Table 6: Quantitative results on the LPBA dataset regarding the impact of pyramid layer numbers.

| layer $l$ | DSC(%) ↑ | %$|J_\phi| \leq 0$ ↓ | HD95 ↓ | ASSD ↓ |
|---|---|---|---|---|
| 1 | 61.2±4.6 | <1.2 % | 6.947±0.763 | 2.20±0.296 |
| 2 | 69.0±2.7 | <0.098% | 6.217±0.632 | 1.79±0.193 |
| 3 | 72.3±2.0 | <0.14% | 5.672±0.552 | 1.59±0.139 |
| 4 | 73.4±1.4 | <0.12% | 5.44±0.47 | 1.53±0.10 |
| 5 | 73.5±1.3 | <0.13% | 5.389±0.439 | 1.51±0.095 |
| 6 | **73.6±1.2** | <0.11% | **5.381±0.408** | **1.51±0.084** |

Table 7: Quantitative results on the Mindboggle dataset regarding the impact of pyramid layer numbers.

| layer $l$ | DSC(%) ↑ | %$|J_\phi| \leq 0$ ↓ | HD95 ↓ | ASSD ↓ |
|---|---|---|---|---|
| 1 | 49.2±2.9 | <2.0% | 6.77±0.58 | 1.79±0.21 |
| 2 | 60.4±2.1 | <0.13% | 6.22±0.54 | 1.45±0.16 |
| 3 | 63.8±1.5 | <0.35% | 5.67±0.47 | 1.28±0.12 |
| 4 | 64.6±1.3 | <0.34% | 5.39±0.40 | 1.21±0.09 |
| 5 | 64.5±1.3 | <0.35% | 5.42±0.40 | 1.21±0.09 |
| 6 | 64.6±1.3 | <0.35% | 5.38±0.39 | 1.21±0.09 |

This section investigates the impact of the number of pyramid layers on registration results by training networks with varying layers on the LPBA and Mindboggle datasets, as shown in Tables 6 and 7. These tables present quantitative results corresponding to changes in the number of pyramid layers for both datasets. Notably, the size of the $l$-th layer feature maps, which matches the original image dimensions of 160×192×160, reduces to $(1/2)^3$ of the previous layer's size with each additional layer, resulting in a feature map size of 5×6×5 for the sixth layer. The data indicate that as the number of pyramid layers increases, registration accuracy improves; however, beyond a certain number of layers, the accuracy does not significantly increase and may even fluctuate.

7) Analysis of the Ablation Experiment

This section further investigates the impact of gradients and differences on the network models using both datasets, with results presented in Tables 8 and 9. The overall performance of the two variants demonstrates that the inclusion of the differential homeomorphism layer enhances registration accuracy and produces more reasonable deformations. Variants that do not incorporate gradients or differences performed the worst across both datasets. Specifically, in the LPBA dataset, the variant that uses only differences exhibits lower DSC, higher Hd95, higher ASSD, and higher Jacobian values compared to the variant that uses only gradients. However, in the Mindboggle dataset, it shows higher DSC, higher Hd95, higher ASSD, and higher Jacobian. Interestingly, in the LPBA dataset, the version of MsMorph_diff that incorporates only gradients achieved the best performance, while a similar phenomenon was not observed in the Mindboggle dataset, suggesting that combining both modules is stronger generalization ability than using either one individually.

Considering that our network structure is not overly complex, adjustments can be made to adapt it to other applications as needed.

Table 8: Ablation experiments on the LPBA dataset, where D and G represent the differences and gradient, respectively.

| Model | D | G | DSC(%) ↑ | $\%\|J_\phi\| \leq 0$ ↓ | HD95 ↓ | ASSD ↓ |
|---|---|---|---|---|---|---|
| MsMorph | | | 72.4±1.6 | <0.08% | 5.57±0.49 | 1.58±0.11 |
| | | ✓ | 73.3±1.5 | <0.1% | 5.44±0.46 | 1.53±0.10 |
| | ✓ | | 73.0±1.6 | <0.1% | 5.49±0.49 | 1.54±0.11 |
| | ✓ | ✓ | 73.4±1.4 | <0.1% | 5.44±0.47 | 1.53±0.10 |
| MsMorph_diff | | | 72.5±1.6 | <0.00004% | 5.56±0.47 | 1.57±0.11 |
| | | ✓ | 73.5±1.5 | <0.00002% | 5.42±0.46 | 1.52±0.10 |
| | ✓ | | 73.0±1.6 | <0.00005% | 5.52±0.52 | 1.55±0.12 |
| | ✓ | ✓ | 73.4±1.5 | <0.00009% | 5.44±0.46 | 1.52±0.10 |

Table 9: Ablation experiments on the Mindboggle dataset, where D and G represent the differences and gradient, respectively.

| Model | D | G | DSC(%) ↑ | $\%\|J_\phi\| \leq 0$ ↓ | HD95 ↓ | ASSD ↓ |
|---|---|---|---|---|---|---|
| MsMorph | | | 63.9±1.4 | <0.3% | 5.50±0.44 | 1.24±0.11 |
| | | ✓ | 63.9±1.3 | <0.3% | 5.39±0.39 | 1.22±0.09 |
| | ✓ | | 64.4±1.3 | <0.3% | 5.42±0.39 | 1.22±0.09 |
| | ✓ | ✓ | 64.6±1.3 | <0.3% | 5.39±0.40 | 1.21±0.09 |
| MsMorph_diff | | | 64.0±1.4 | <0.00002% | 5.47±0.42 | 1.24±0.10 |
| | | ✓ | 64.1±1.3 | <0.000007% | 5.38±0.39 | 1.21±0.09 |
| | ✓ | | 64.5±1.3 | <0.00003% | 5.42±0.41 | 1.22±0.09 |
| | ✓ | ✓ | 64.8±1.3 | <0.004% | 5.39±0.39 | 1.20±0.09 |

## 8) Limitations

This study has thoroughly considered the differences and complexities of human brain structures during the registration process, achieving accurate alignment among brain images from different individuals. However, there are some limitations. Firstly, the work was conducted exclusively on brain MRI images. Secondly, instead of conducting an extensive grid search to identify the optimal hyperparameters for baseline methods, values were determined based on empirical observations or recommendations from the original papers. Thirdly, the focus was primarily on the methodology, with limited investigation into the effects of the loss function and other hyperparameters. In the future, we aim to refine this method further and explore its application to other imaging modalities, such as CT and PET scans, to enhance generalizability.

## V. CONCLUSION

We propose a deep learning framework for unsupervised medical image registration that exhibits strong interpretability. By leveraging the scale invariance of images and the feature similarity between registered pairs, this approach optimizes the parameters for registration. It combines the advantages of traditional pyramid structures with decomposed deformation fields, modeling the differences between the gradients of image pairs to direct the network's focus toward regions requiring deformation, and sufficiently imitates the process of manual registration, constantly comparing and distorting images. This process encourages the model to produce results that increasingly approximate the true deformation field. In addition, the introduction of gradient information enhances the network's ability to perceive voxel-level changes, promotes the extraction of structural features, and improves the fitting of network parameters. The proposed method aligns more closely with human cognitive processes, thereby enhancing the interpretability of the model. Experimental results demonstrate the effectiveness of the proposed method.

## Acknowledgments

**Funding**: This work was supported by the National Natural Science Foundation of China [No.62476255 and No. 62303427], the Young Teacher Foundation of Henan Province [No.2021GGJS093], the Key Science and Technology Program of Henan Province [No.242102211058 and No.242102211018], the Key Science Research Project of Colleges and Universities in Henan Province of China [No.25A520003], and the Young Backbone Teacher Program of Zhengzhou University of Light Industry.